\documentclass[sigconf]{acmart}
\usepackage[nolist]{acronym}
\usepackage{listings}
\usepackage{pifont} % checkmark symbols
\usepackage{subfig}
\usepackage{multirow}

\newcommand{\cmark}{\ding{51}}%check mak yes
\newcommand{\xmark}{\ding{55}}%check mark no

\def\revcolor{black}
\newcommand{\rev}[1]{{\color{\revcolor}#1}}

\AtBeginDocument{%
  \providecommand\BibTeX{{%
    \normalfont B\kern-0.5em{\scshape i\kern-0.25em b}\kern-0.8em\TeX}}}

\copyrightyear{2023}
\acmYear{2023}
\setcopyright{acmlicensed}\acmConference[HIP '23]{7th International Workshop on Historical Document Imaging and Processing}{August 25--26, 2023}{San Jose, CA, USA}
\acmBooktitle{7th International Workshop on Historical Document Imaging and Processing (HIP '23), August 25--26, 2023, San Jose, CA, USA}
\acmPrice{15.00}
\acmDOI{10.1145/3604951.3605511}
\acmISBN{979-8-4007-0841-1/23/08}

\begin{document}

\title{DIVA-DAF: A Deep Learning Framework for Historical Document Image Analysis}

\author{Lars Vögtlin}
\email{lars.voegtlin@unifr.ch}
\affiliation{%
  \institution{University of Fribourg}
  \streetaddress{Bd de Pérolles 90}
  \city{Fribourg}
  \country{Switzerland}
  \postcode{1700}
}

\author{Anna Scius-Bertrand}
\email{anna.scius-bertrand@unifr.ch}
\affiliation{%
  \institution{University of Fribourg}
  \streetaddress{Bd de Pérolles 90}
  \city{Fribourg}
  \country{Switzerland}
  \postcode{1700}
}

\author{Paul Maergner}
\email{paul.maergner@unifr.ch}
\affiliation{%
  \institution{University of Fribourg}
  \streetaddress{Bd de Pérolles 90}
  \city{Fribourg}
  \country{Switzerland}
  \postcode{1700}
}

\author{Andreas Fischer}
\email{andreas.fischer@unifr.ch}
\affiliation{%
  \institution{University of Fribourg}
  \streetaddress{Bd de Pérolles 90}
  \city{Fribourg}
  \country{Switzerland}
  \postcode{1700}
}

\author{Rolf Ingold}
\email{rolf.ingold@unifr.ch}
\affiliation{%
  \institution{University of Fribourg}
  \streetaddress{Bd de Pérolles 90}
  \city{Fribourg}
  \country{Switzerland}
  \postcode{1700}
}

\renewcommand{\shortauthors}{Vögtlin et al.}

\begin{abstract}
  Deep learning methods have shown strong performance in solving tasks for historical document image analysis. However, despite current libraries and frameworks, programming an experiment or a set of experiments and executing them can be time-consuming. This is why we propose an open-source deep learning framework, DIVA-DAF, which is based on PyTorch Lightning and specifically designed for historical document analysis. Pre-implemented tasks such as segmentation and classification can be easily used or customized. It is also easy to create one's own tasks with the benefit of powerful modules for loading data, even large data sets, and different forms of ground truth. The applications conducted have demonstrated time savings for the programming of a document analysis task, as well as for different scenarios such as pre-training or changing the architecture. Thanks to its data module, the framework also allows to reduce the time of model training significantly. 
\end{abstract}






\received{12 May 2023}

\maketitle

% ---------------------------------------------------------------
% SECTIONS
% ---------------------------------------------------------------
\begin{acronym}[UMLX]
 \acro{HDIA}{Historical Document Image Analysis}
 \acro{DIA}{Document Image Analysis}
 \acro{ML}{Machine Learning}
 \acro{CUDA}{NVIDIA Compute Unified Device Architecture}
 \acro{CuDNN}{Deep Neural Network library}
 \acro{CNN}{Convolutional Neural Network}
 \acro{CLI}{Command Line Interface}
 \acro{DL}{Deep Learning}
 \acro{CV}{Computer Vision}
 \acro{NLP}{Natural Language Processing}
 \acro{PR}{Pattern Recognition}
 \acro{mIoU}{mean Intersection over Union}
 \acro{PL}{PyTorch-Lightning}
 \acro{EECVF}{End-to-End CV Framework}
 \acro{CI}{Continuous integration}
\end{acronym} 

\newpage
\section{Introduction}

% Analyser automatiquement les collections de documents historiques contribuent à préserver le patrimoine culturelle. Même si de grandes avancées ont été effectués ces dernière années, ce domaine reste un challenge (\cite{fischer20handwritten}). Au vue de la grande variabilité des collections aussi bien sur la forme que le contenu, les methodes de deep learning sont prometeuses. 

Automatically analyzing collections of historical documents provides strong support for the preservation of our cultural heritage. Although great progress has been made in recent years, this field of research remains a difficult challenge, especially due to the high variability of document collections both in terms of form and content~\cite{fischer20handwritten}. Deep learning methods have shown a strong potential for historical document image analysis, achieving state-of-the-art results for different tasks ranging from layout analysis over handwriting recognition to information retrieval.

% Deep learning methods have played a significant role in the last decade in increasing the performance of Computer Vision tasks. Même si les progrés sont normes dans le champs de l'analyse de documents, certains domaines restent un défis tels que l'analyse de document historiques (histoc book). Pourtant analyser automatiquement les larges et diverses collections de manuscrits anciens contribuent à preserver le patrimoine culturelle. Au vue de la grande variabilité des collections aussi bien sur la forme que le contenu, les methodes de deep learning sont prometeuses. 

% Un des défis majeurs de l'analyse de documents historiques résident dans leur grandes variabilité de la forme et du contenu, la présence de dégradation ainsi qu'un accès souvent difficile à des données d'entraînement. 

% Deep learning had a huge influence on the image and especially the document image analysis community.
%vIt replaced the art of extracting meaningful features from a highly complex input as is usual in document image analysis and, in the meantime, surpassed human-level performance.
% This makes it easier for non-export to enter the field and participate.
%With the help of deep learning, there was a shift from fast learning and slow prediction methods to slow training and fast prediction methods.
% This reduced the time to predict a large number of samples and makes the method much more scalable compared to traditional document image analysis methods.

% what are the needs for a scientific experiment 

% *Existing tools:
%     - library: very modular but
%         - limit (2-3)

In order to use deep learning more quickly and efficiently, several libraries were created (e.g., Tensorflow, Keras, PyTorch, ...). %, Tensorflow, Caffee).
Such libraries are built in a very open and general fashion to allow good programmers to take advantage of the full potential of this new technology.
However, due to their generality, these libraries have a steep learning curve.
Additionally, the user usually has to take care of the whole hardware orchestration, such as moving data to the GPU or aggregating certain information across a number of devices.
Therefore, software frameworks built on top of the general deep learning libraries may significantly facilitate the application of deep learning to specific types of data and tasks.

%Using a framework can prevent managing the hardware orchestration. Besides, the framework brings a lot of time-saving, especially around the bolding plate. But using a framework can lead to a lack of flexibility. 

%     - FW: not modular by design: 
%         - General explication of the pb non-modularity design, too complex, time-consuming 
% General frameworks solve certain problems of these libraries by taking away the hardware responsibility from the user and flattening the learning curve. These frameworks are, most of the time, based on a deep learning library but are no longer built in a modular fashion.
% This non-modularity can make these frameworks time-consuming and very complex to introduce new features for non-expert users of these frameworks.

% This causes problems while introducing new features as the interoperability of different framework parts gets interfered with. The consequence is a high code complexity, and it is very time-consuming to extend the feature set of such frameworks. Hence, most frameworks have a low learning curve if you don't want to adapt functionality but are very time-consuming when introducing new methods. 

In the context of historical document analysis, a deep learning framework must be able to deal with large images and support different ground truth formats. Furthermore, one of the biggest challenges for historical documents is the lack of training data. It is often necessary to use different learning strategies such as transfer learning, self-learning, or data generation. Conducting scientific experiments with advanced learning mechanisms requires a high flexibility from the deep learning framework, to be able to iteratively test different parameters and configurations without having to reprogram the whole experiment. Moreover, a prerequisite of any experiment is its reproducibility. 

% Dans le cadre de l'analse de document historiques, un framework doit être en mesure de charger des images de grandes tailles et de prendre en charge différents formats de ground truth. De plus, un des défis majeurs de l'analyse de documents historiques est l'absence de données d'entrainement. Il est souvent nécessaire de pouvoir utiliser différentes stratégies d'apprentissage tel que le transfert learning, le self learning ou encore la génération de données. 

%Conducting scientific experiments requires high flexibility to be able to iteratively test different parameters and configurations without having to reprogram the whole experiment. Moreover, a prerequisite of any experiment is its reproducibility. 

% Conduire des expériences scientifiques requièrent de pour introduire de nouvelles fonctionnalities. De plus cela nécessitent une grande flexibilité pour pouvoir tester de manière itérative des paramètres et différentes configurations sans avoir à devoir reprogrammer l'ensemble de l'expérimentation. De plus, un pré-requis de toute expériences est sa reproductibility. 

In this paper, we introduce a new deep learning framework, DIVA-DAF\footnote{https://github.com/DIVA-DIA/DIVA-DAF}, which is specifically designed for conducting experiments in the domain of image analysis for historical documents. The framework is based on PyTorch Lightning, which allows us to benefit from its flexibility and hardware management. We have added features to conduct document analysis experiments with a view to reproducibility, maintainability, efficiency, and increased flexibility. 
These features allow, among others: rapid prototyping, faster runtime, use of transfer learning and self-learning, simplified change or swap of network parts, adding own code with unit tests, using external libraries, simplified change (without reprogramming) of networks, tasks, and datasets in an experiment, and keeping track of previous experiments. These features make our framework a powerful tool to conduct experiments in the context of historical document image analysis.

% C'est pourquoi nous proposons un nouveau framework de deepl learning spécialement conçu pour la conduite d'experience en analyse d'image de documents historiques. Ce framework est basé sur Pytorch lighting qui permet de bénéficier de sa flexibilité et la gestion du hardware. Nous avons ajouter des features pour conduire des expérience analyse de document tel que la reproductibility, maintability, l'efficency ou encore d'avantage de flexibility. Ces nouveaux features permettent entre autre : rapid prototype, faster runing time, easily change or intervertir part from one or several network, use transfert learning, self-learning, without re-programming change a task or a network or a dataset in an experimentation, add your own code with a test, used a external libraries, easily keep track of your previous experimentation. These features make our framework a powerful tool to conduct experiments on historical documents analysis.

DIVA-DAF supports all types of deep neural networks and learning strategies, including supervised, semi-supervised, and unsupervised learning. Therefore, all typical document image analysis tasks can be implemented with our framework, such as layout analysis, line segmentation, keyword spotting, transcription alignment, handwriting recognition, etc. At the moment, \rev{two} tasks are already implemented and readily available: segmentation and classification. For example, to perform semantic segmentation for layout elements in a historical document, it is sufficient to indicate the data and ground truth folders in a configuration file. In this configuration file, it is also possible to change the network model, loss function, optimizer, and evaluation metrics, among others. In less than 5 minutes, a custom experiment can be configured and started.

In the remainder of this paper, we provide an overview of related work in Section~2. Then, we introduce the DIVA-DAF framework and describe its features in Section~3. Afterwards, we present a case study using the framework in Section~4. Finally, we provide some conclusions and an outlook to future work in Section~5.

\section{Related Work}
% In our literature research, we came across different existing frameworks for \ac{DL} with which we share the general motivation as well as ideas.

% We categorized the frameworks by their main focus into different groups.

% data and or code management 
% controlDataVersionControl2017
%Data Version Control~\cite{controlDataVersionControl2017} takes care of managing your code and data in an git-like fashion.

% Data Version Control~\cite{controlDataVersionControl2022} takes care of managing \ac{ML} projects, including code and data.
% This is achieved with a git-like \ac{CLI}.

% % track code, experiments, and results
% Comet~\cite{CometSuperchargeMachine2022}, Neptune~\cite{NeptuneAiMetadata2022}, and Weights and Biases~\cite{biewaldExperimentTrackingWeights2020} take care of tracking your code, experiments, and results as well as visualizing them.
% They work with a large amount of different machine-learning libraries.

% We have end-to-end framworks in CV
In our literature research, we came across different end-to-end frameworks for Computer Vision and Document Image Analysis with which we share the general motivation as well as ideas.

%TODO camera-ready: extend the description of Transkribus and eScriptorium

%Transkribus
Transkribus is a well-known platform designed to automatically transcribe historical documents. The platform offers the possibility to train a model with specific data and also provides trained text recognition models ready to be used.
%based on PyLaia~\cite{puigcerver2018pylaia}.
An interface is available. But the platform is not open source and provides only a restricted number of document analysis tasks. 

% Escriptorium
Another well-known platform for automatic transcription of historical documents is eScriptorium~\cite{kiessling2019escriptorium}, which is open source. The text recognition system is based on Kraken~\cite{kiessling2019kraken}. Other models can be integrated into the platform. But as Transkribus, eScriptorium is also limited in the number of document analysis tasks covered.

% Chainer
Chainer~\cite{tokuiChainerDeepLearning2019} provides a wide range of \ac{DL} models for researchers in a flexible and intuitive fashion. 
The framework's focus is high-performance and distributed training, which is achieved with the help of standard Python libraries.
But the framework no longer gets updates (last release June 2022), it misses an easy definition of an experiment, and networks can not be loaded in parts.
Additionally, as it is based on plain NumPy, there is a lack of compatibility with other \ac{DL}-frameworks.

% EECVF
Orhei et al.~\cite{orheiEndToEndComputerVision2021} introduced an \ac{EECVF} to tackle the problem of creating a true end-to-end \ac{ML} platform that allows combining \ac{DL} approaches together with classical \ac{PR} methods.
Their platform is constructed to be easily usable for research and educational purposes.
It uses multiple configuration files to define the behavior of the different parts without the need to write code.
The biggest problem with this framework is that it is no longer available, does not support different hardware accelerators, has limited logging, and no flexible loading of a network's weights.

% VISSL
Goyal et al.~\cite{goyal2021vissl} from Meta Research created a PyTorch-based framework named VISSL for self- and unsupervised pretraining of neural networks for natural images.
The main idea of this framework is to provide a fast and easy way to pretrain neural networks with natural images in a self-supervised fashion with the help of a configuration system.
Additionally, hardware acceleration with GPUs, logging with Tensorboard, and a large variety of preimplemented methods and datasets are a part of it.
This project also has some issues: It is no longer maintained (last release November 2021), it is not possible to fine-tune a network on some final \rev{tasks}, logging is limited to Tensorboard, and introducing new networks, functionalities, or other parts is very tedious work.
% no longer maintained; you need a separate framework to fine-tune the network; logging restricted to tensorboard

% DeepDIVA
DeepDIVA was introduced by Alberti et al.~\cite{albertiDeepDIVAHighlyFunctionalPython2018a,albertiImprovingReproducibleDeep2019} as an out-of-the-box deep learning framework for \ac{CV}.
The focus of the framework was to provide reproducible experiments that the user can redefine based on existing networks, datasets, and parts, but they have the possibility to add their own.
It also provides \rev{e.g.,} hardware acceleration with GPUs, logging with Tensorboard, reproducibility by versioning the code, and different visualizations of the data.
However, there are several problems with this framework: Introducing custom network parts is difficult as the framework is not built in a modular fashion, all parameters are handed over via \ac{CLI}, which makes it difficult to read, the weight loading functionalities are limited, and logging is only provided for Tensorboard.

\rev{A library approach was taken by Shen et al.~\cite{shen2021layoutparser} with their LayoutParser. 
Their goal is to simplify the construction of \acl{DL} workflows within the \ac{DIA} domain. 
They additionally provided a platform to share models, code, and weights. 
This project seems no longer actively supported as their last change is from August 2022.}

% PyTorch-Lightning
Falcon et al.~\cite{falconPyTorchLightning2019} started 2019 the modular PyTorch-based general \ac{DL} framework \acf{PL}.
It focuses on rapid prototyping, wide hardware integration, and maximal flexibility.
Additionally, it takes advantage of the large ecosystem, providing implementations for metrics, models, data modules, and other state-of-the-art functionalities.
As the framework is not focusing on \ac{CV} or \ac{DIA}, it lacks the support to handle large images and does not provide an easy way to create an experimental setup.

To have an overview of the lack of modularity and flexibility of the different frameworks, see Table~\ref{tab:realted_work_modularity}.
With DIVA-DAF, we added all this modularity and flexibility of the different categories that are relevant for deep learning experiments in the context of historical documents. They are further introduced in the following section.

% TODO: Should we add what DIVA-DAF does or is that already but in the introduction?
% TODO: Do we also add the non-end-to-end framworks to the mix? iF yes how?

\begin{table*}[h]
  \caption{Modularity of Different Frameworks in Different Categories}
  \label{tab:realted_work_modularity}
  \begin{tabular}{ccccccccc|c}
    \toprule
    Name                & DeepDIVA  & VISSL     & Chainer   & EECVF     & PL                & DIVA-DAF \\
    \midrule
    Input               & \cmark    & \cmark    & \cmark    & \cmark    & \cmark            & \cmark \\
    Output              & \cmark    & \xmark    & \cmark    & (\xmark)  & \cmark            & \cmark \\
    Hyper-parameters    & \cmark    & \cmark    & \cmark    & \cmark    & \cmark            & \cmark \\
    Network             & \xmark    & \xmark    & \xmark    & (\xmark)  & \xmark            & \cmark \\
    Monitoring          & \xmark    & \cmark    & \cmark    & (\xmark)  & \cmark            & \cmark \\
    Evaluation          & \cmark    & \xmark    & \cmark    & \cmark    & \cmark            & \cmark \\
    Reproducibility     & \cmark    & \cmark    & \cmark    & \cmark    & \cmark            & \cmark \\
    Experimental setup  & \xmark    & \cmark    & \xmark    & \cmark    & \xmark            & \cmark \\
    \bottomrule
  \end{tabular}
\end{table*}

\section{DIVA-DAF - Document Analysis Framework}
To be able to conduct scientific experiments on document image analysis, our framework has the following characteristics: flexibility, efficiency, reproducibility, and maintainability. 
The framework is mainly designed for experienced programmers in document image analysis, but thanks to a configuration system, non-experts with few programming skills can also create, launch, and interpret experiments. 
In this section, we introduce the deep learning framework DIVA-DAF and explain its main attributes. 

\subsection{Flexibility}
\begin{figure*}[h!]
    \centering
    \includegraphics[width=\textwidth]{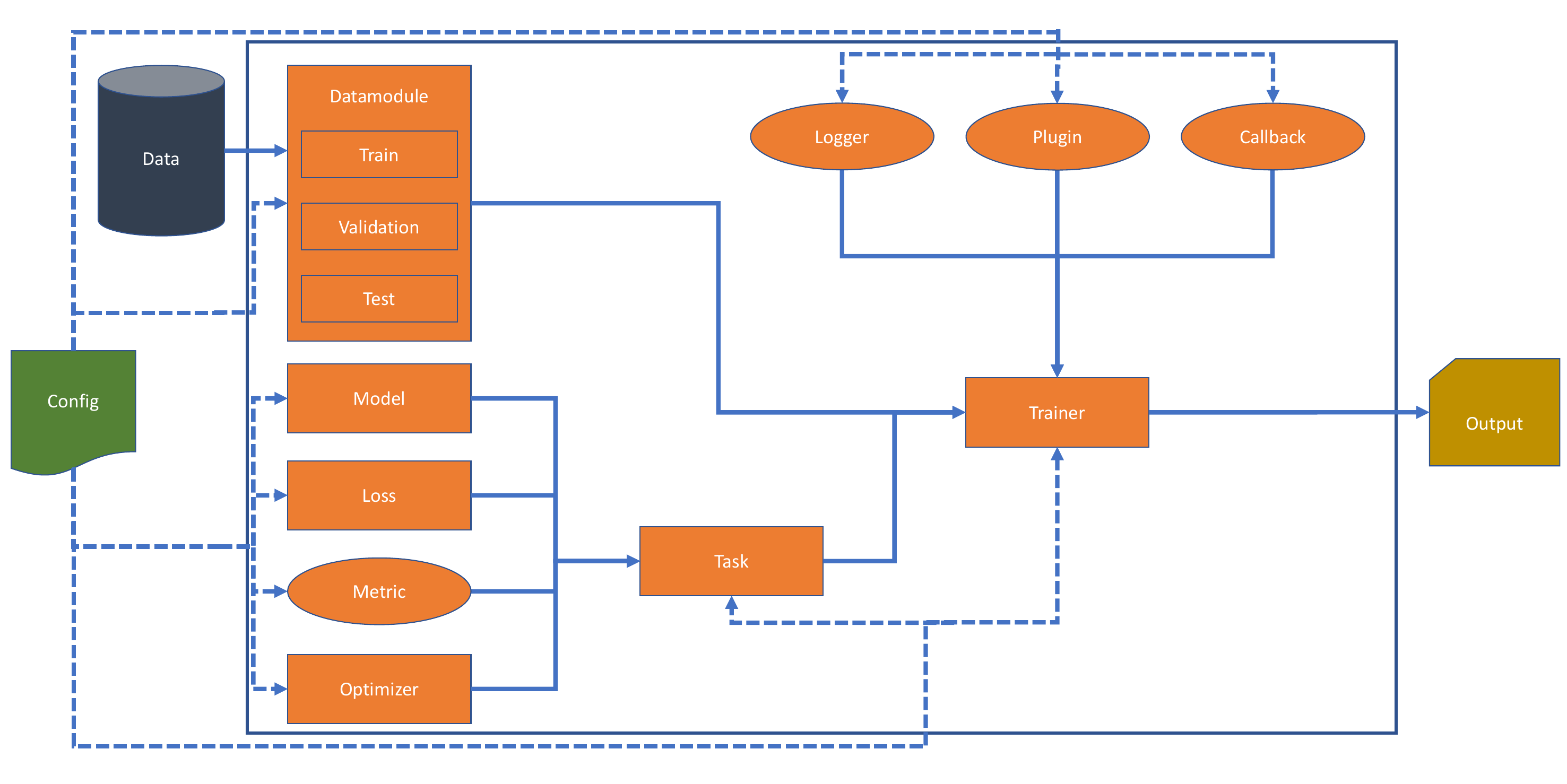}
    \caption{The module schema of DIVA-DAF. Rectangles represent required components, ovals represent optional components, and green is the configuration.}
    \label{fig:framework_schema}
\end{figure*}

% make it more a sequential story describing each module, that its interchangeable and why do we want to be able to easily exchange it
The framework consists of several components following an object-oriented programming paradigm. 
% The framework's components are based on PyTorch Lightning~\cite{falconPyTorchLightning2019}, a PyTorch research framework that focuses on the rapid development of research experiments. 
The general architecture is presented in Figure~\ref{fig:framework_schema}.

Each component is independent and can be easily changed.
Like classical frameworks, data is loaded into a data module. 
But unlike other platforms, the data module manages the different datasets needed for each stage (training, validation, testing, prediction) in a modular way. 
Furthermore, it calculates data statistics and defines special data handling, such as data augmentation and transformations. 
Besides, the data module is able to load large images thanks to two strategies: scaling down the image (adapting also the ground truth) and patch-based approaches.  
Different ground truth formats can already be easily handled: images (color encoded, index encoded, channel encoded) \rev{and} classes based on folder structures.

% own representation gt -> classe system and give it back
% Four main components of our framework are (1) data module, (2) model, (3) task, and (4) trainer (see Fig.~\ref{fig:framework_schema}). 

Compared to PyTorch Lightning, the LightningModule is separated into two components: the model, which defines the neural network architecture, and the task, which describes the task to be solved.
By defining these components independently, the same task can be solved using different models, or the same model can be used to solve different tasks.

% TODO caption text

% explanation of the modular parts
%The data module handles the data and creates the different datasets needed for each stage (training, validation, testing, prediction). 
%Furthermore, it calculates data statistics and defines special data handling, such as data augmentation and transformations.

The model specifies the neural network architecture by defining the backbone and the header. 
The backbone acts as the encoder part of the network, and the header as the classifier. 
By defining these two parts separately, the framework can save them independently and combine them with other backbones or headers.

The task defines the workflow during training, validation, testing, and prediction. 
It requires four inputs: a loss function, an optimizer, metrics, and a model. 
All these components can be easily customized by the user. 
Also, it produces the test output and provides the needed method to bring the network output into a specific loss or metric format. 

The trainer connects the different components of our framework and runs them. 
It executes the different stages - training, validation, testing, and prediction - and runs the neural network. 
It is responsible for initializing the different hardware devices and moving data and models to the correct device. 
The default implementation of PyTorch Lightning is used, but users could also exchange or modify this part if required. 
Trainers are also connected to a logger and a callback component which will be described in subsection \ref{sec:reproductibility}, and a plugin component which enables changing the behavior of the trainer, e.g., custom precision or cluster environment implementations. 

% TODO: Explain config system (whats in the config file and how is it used)

% In addition to the four core modules, the framework contains several smaller components: loss, optimizer, metric, logger, callbacks, modes, and plugins. 
% Loss, optimizer, and metric are a subpart of the task module. 
% The logger provides the possibility to use the user's favorite logging system. 
% With plugins and callbacks, which PyTorch Lightning provides, the user can control the behavior of devices, stage loops, and other parts of the framework.

\subsection{Efficiency}
% The framework assures efficiency at three levels: implementation, setup, and running time. 

% TODO improve intro sentece
The first obstacle to using a new framework is the difficulty related to its installation. 

% add intro sentence, efficency for implementation, setup, running

% Implementation time (time to create a new experiment or introduce new functionality)
% For a researcher, it is important to check in a short amount of time if an idea is worth further investigation or not.
% So the time to install a framework and set up an experiment should be minimized.

DIVA-DAF is easy to set up: The user clones the code from the GitHub repository (\url{https://github.com/DIVA-DIA/DIVA-DAF}) and creates a new Python environment (e.g., Anaconda\cite{anaconda} environment) based on the shipped requirement file.
This requirement contains all dependencies with the corresponding versions to run the framework.
If a user wants to run the framework with GPU, TPU, HPU, MPS, or IPU support, the appropriate drivers and the correct PyTorch support packages (CUDA, ROCm, etc.) must be installed.

% Implemented tasks

Thanks to the modular structure of the framework, users can do rapid prototyping.
The user can easily combine existing modules into a new experiment. 
The hyperparameters of an existing one can be changed without writing a single line of code.
To create or introduce new modules and swap them out with existing ones, just a few lines of code are needed, as the framework provides templates for the different modules. 
An example is given in Section~\ref{sec:application}.

% Run time (efficient hardware usage, hence, faster run times)
By using the full potential of the underlying PyTorch Lightning framework, DIVA-DAF takes full advantage of any hardware provided by the host system, optimized data-loading strategies, and distributed computing.
% precise what is a gain of running time from PL et what we add
With an efficient implementation of our datasets, we were able to further reduce the runtime of the experiments.

\subsection{Reproducibility}
\label{sec:reproductibility}
% Configuration system (store everything at the beginning of a run)
Another important attribute of a framework is its ability to create reproducible research experiments.
% 1 introduction comment : save + logging
% save parameter: 1 sentence to present config file: we have a config file to store all importante parameter to produce. The content is : ..., ... ,.. ,.. 
To make each experiment as reproducible as possible, we store the configuration file alongside the results and network weights of each run in its output directory.
It also saves the seed used to initialize the pseudo-random generators used during training to initialize the neural network weights and other environmental information. 
% We also sotre the git commit number to have information about the version of the code used for the experiment.
Using the configuration file with this seed, anyone can quickly reproduce published results with this framework.

% Logging
To keep track of experiments, we use the logging functionality of PyTorch Lightning.
They provide the most common loggers like Weights and Biases~\cite{biewaldExperimentTrackingWeights2020} or Tensorboard.
The framework allows using multiple loggers simultaneously.
% a CSV logger is always running (implement that?)
Besides using a cloud-based logger, like Weights and Biases, the user can also use a local logger like the CSV logger.
The CSV logger writes all the logging information into the local experiment folder for later use.
% the user can implement or add new loggers the same way as other modules
Logging is not limited to scalar data like metrics or loss information.
It is also possible to log figures, images, or histograms, but each logger needs to do this individually.
As with all the other modules of our framework, users can also implement their own logger or adapt an existing one.

\subsection{Maintainability}
% introduction sentence how the section is structured and what we are talking about and why
Maintainability is a key concept to create a long-lasting framework. 
DIVA-DAF provides two maintainability factors: injecting external code and intercompatibility. 
% Inject external code
% - Callbacks
Users can inject code via callbacks provided by the underlying \ac{PL} framework.
Callbacks hook into predefined methods and can be used at each stage of the experiment.
The main advantage of callbacks is that the user does not have to change the core code of the framework.
Hence, it provides extendability without the cost of damaging the integrity of the system.

% - CI (Code quality and Code integrity checks)
Additionally, DIVA-DAF uses GitHub actions to provide \ac{CI}.
For every change in the framework, the different modules get extensively tested with unit tests, and the code quality (duplication, bugs, complexity) gets checked.
This ensures a good code base and gives the user the possibility to check if these changes break anything in the framework.
% Intercompatibility (can work with different other libraries [table 2])

Thanks to its modularity and compatibility with the PyTorch ecosystem, any module from PyTorch-based libraries can be integrated into DIVA-DAF.

\section{Applications}
\label{sec:application}

In this section, we compare the programming time and execution time of a document analysis experiment using DIVA-DAF and \acf{PL}. 

\subsection{Methods}
To compare the programming time, a baseline experiment was performed and then broken down into three scenarios. 

The baseline experimentation consists of semantic segmentation for layout elements in historical documents. 
Each pixel of an input image gets assigned one of the predefined classes. 
An experienced \ac{PL} programmer timed each of the development steps using the two different frameworks.

Scenario 1 - S1: Pre-training / transfer learning: When the volume of training data is too small, pre-training or transfer learning can improve the performance of the network. 
In this case, the first \textit{n} layers of an already pre-trained U-Net~\cite{ronnebergerUnetConvolutionalNetworks2015} (here n=3) were loaded into a randomly initialized U-Net, and afterward fine-tuned on the additional data.

Scenario 2 - S2: Comparison of the network architecture: Often in research, it is necessary to compare several networks for the same task on identical data. 
Here the task is to replace the U-Net with DeepLabV3~\cite{chen2017rethinking}, an architecture already implemented in Torchvision.

Scenario 3 - S3: Visual control during training: training a network can lead to a ``black box'' effect. 
Visualizing an intermediate result during training can be useful for understanding the behavior of the network. 
The task here is to be able to save n images randomly from the validation set (here 1 image).

To compare the execution time, we replicated the experiment by Studer et al.~\cite{studerComprehensiveStudyImageNet2019a} (same network running on the same hardware), with the difference that we performed the experiment with DIVA-DAF.

\subsection{Data}
\begin{figure*}[ht] 
    \centering
    
    \subfloat[CB55, p. 25r]{
    \includegraphics[width=\columnwidth]{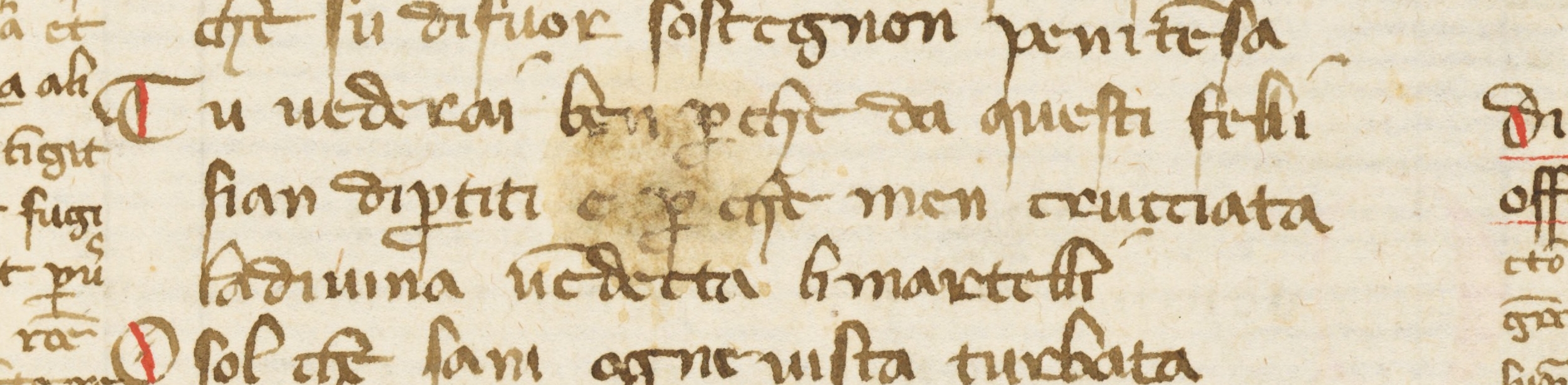}\label{fig:dataset_cb55_A}}
    \hfill
    \subfloat[CB55, p. 5v]{
    \includegraphics[width=\columnwidth]{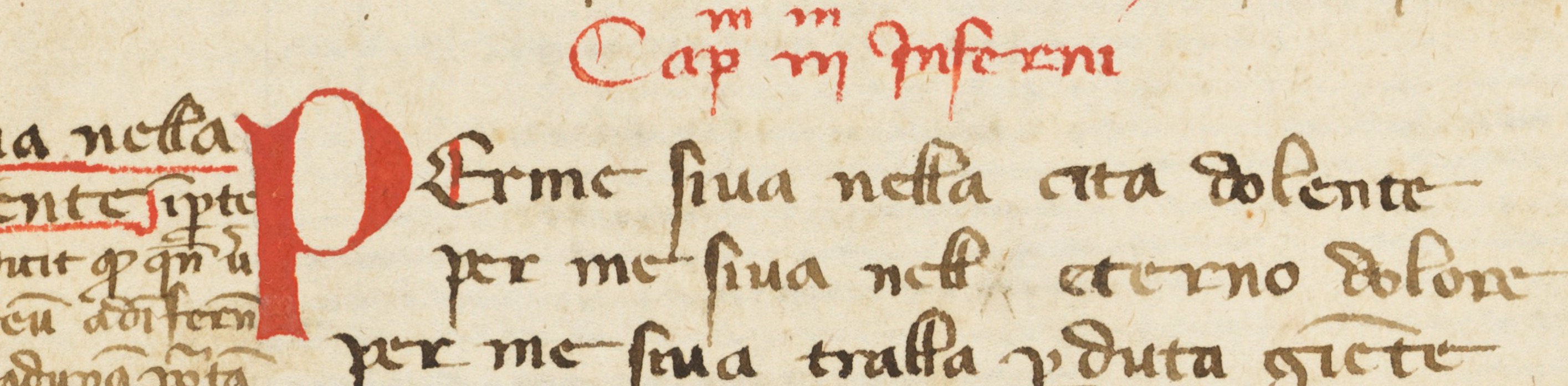}\label{fig:dataset_cb55_B}}
    
    \caption{
    Sample pages of the medieval manuscripts Codex Bodmer 55 of DIVA-HisDB.
    }
    \label{fig:dataset}
\end{figure*}

% U-Net Semantic segmentation DIVAHisDB
The dataset is the Codex Bodmer 55 of the DIVA-HisDB~\cite{simistiraDIVAHisDBPreciselyAnnotated2016} dataset (see Fig.~\ref{fig:dataset}). 
The dataset contains 20 pages for training, 10 pages for validation, and 10 pages for testing. Each page has a dimension of 4872$\times$6496 pixels with a resolution of 600 dpi.
Each pixel belongs to one of 8 classes (background, main text body, decoration, comment, main text body + comment, main text body + decoration, comment + decoration,  main text body + decoration + comment).

% Example Image CB55

\subsection{Results}
% Make a step by step description of how we implemented the PL and the DAF part
The baseline implementation is split into three parts: data loading, network, and execution.
Data loading includes the dataset, calculating statistics on the training data, normalizing the data, and creating a data module.
The network part is just the implementation of the network and its behavior in the different stages.
Last, in the execution part, we combine the two parts from above into a runnable experiment.
For the time we used to implement this and its code duplication, see Table~\ref{tab:application_implementation}.

The time to implement the baseline experiment in \ac{PL} is nearly 15x longer compared to the same experiment in DIVA-DAF.
To load the data, which is the DIVA-HisDB~\cite{simistiraDIVAHisDBPreciselyAnnotated2016} format, it just takes a few lines of YAML (see Listing~\ref{lst:datamodule}).
In our main experiment configuration, we need to specify the data module we want to use (\texttt{\_target\_}), the path to the data (\texttt{data\_dir}), how big our crops should be (\texttt{crop\_size}), and the batch size (\texttt{batch\_size}).

In contrast, in \ac{PL}, we have to implement the whole data loading logic, as well as take care of calculating statistics, multi-device training, and applying transformations.
These parts are very crucial to have a correctly working experiment and so take a lot of time to implement.

\begin{lstlisting}[caption={The config describing the data module},breaklines=true,label={lst:datamodule}]
datamodule:
    _target_: src.datamodules.DivaHisDB.datamodule_cropped.DivaHisDBDataModuleCropped

    data_dir: /net/research-hisdoc/datasets/semantic_segmentation/datasets_cropped/CB55
    crop_size: 256
    batch_size: 16
\end{lstlisting}

The other part that takes more time in plain \ac{PL} is the implementation of the network.
We can take advantage of the U-Net class from the Torchvision library, but we still have to implement the behavior of the network during the training, validation, and testing stages.
In DIVA-DAF, we do not have to do that because the network's behavior is defined in the task and not the network.

To implement the execution part of the scenario, the time difference is not significant.
In \ac{PL}, it takes a few lines of code to create a \texttt{Trainer} object and hand it over to the data and the network.
In DIVA-DAF, we have to adapt the experiment configuration.

\begin{table}[h]
  \caption{Programming time in minutes}
  \label{tab:application_implementation}
  \begin{tabular}{lcccc}
    \toprule
    % Module part & \multicolumn{2}{c}{Programming time [m]} \\
     Tasks & PyTorch L. & DIVA-DAF \\
    \midrule
    Data loading            & 150 & 2 \\
    Network                 & 30  & 5  \\
    Execution               & 10  & 5 \\
    \midrule
    S1. Pre-training        & 15 & 2  \\
    S2. Comparing network   & 20 & 2 \\
    S3. Visualizing         & 25 & 20  \\
    \bottomrule
    % Total                   & 250 & 37 \\
  \end{tabular}
\end{table}

For the first scenario, we can use the preimplemented functionality of DIVA-DAF, where we can define in the configuration the layers of the network we want to load.
The same in \ac{PL} takes more time as you need to filter out the layers we want to use from the checkpoint file and load them into the network.
For an experienced \ac{PL} programmer, this is not a complicated but a time-consuming task.

In the second scenario, we use the Torchvision library again to apply a DeepLabv3 model with a ResNet50~\cite{heDeepResidualLearning2016} backbone to our task.
As the task stays the same, in DIVA-DAF, we have to create a config file (see Listing~\ref{lst:deeplabv3}) for the new network and adapt the experiment.
In \ac{PL}, we can take advantage of the U-Net implementation and copy the code defining the behavior of the network during the different stages.
This creates code duplication, which makes the code harder to maintain and adapt.

\begin{lstlisting}[caption={The config for the DeepLabv3 network with a resnet50 backbone},breaklines=true,label={lst:deeplabv3}]
_target_: torchvision.models.segmentation.deeplabv3_resnet50
num_classes: ${datamodule:num_classes}
\end{lstlisting}

Copying the code in the \ac{PL}, implementation becomes a problem in the third scenario, as we want to change the behavior of the network in the validation stage to save a random image.
We have to copy \rev{the code again} into both network implementations, which creates more code duplication and increases the complexity.
In DIVA-DAF, the code has just to be changed in the task class.
This could, in both cases, also be solved with the help of a callback, which would reduce code duplication but takes our programmer still less time to implement in DIVA-DAF than in \ac{PL}.

For the execution time comparison (see Table~\ref{tab:segmentation_results}), we used the same hardware and hyperparameter as Studer et al.~\cite{studerComprehensiveStudyImageNet2019a}, as well as the SegNet~\cite{badrinarayanan2017segnet} and DeepLapv3~\cite{chen2017rethinking} architecture.
The hardware is a server with 4 $\times$ NVIDIA 1080 GTX with 8GB of GPU memory each, an Intel i7-5960X CPU, and 64 GB of RAM.
The hyperparameters are available here~\footnote{\url{https://bit.ly/2I8c3dX}}.

\begin{table}[h]
    \caption{Results of semantic segmentation on the test set of DIVA-HisDBs CB55. All our networks were trained for 50 epochs. All experiments are conducted on the same hardware.}
    \label{tab:segmentation_results}
    \begin{tabular}{lllrr}
        \hline
        Authors & Year &  Model & Runtime & \acs{mIoU}[\%]\\
        \hline
        % Studer et al.~\cite{studerComprehensiveStudyImageNet2019a} & 30 & SegNet  & $\sim$8h    & 86.90 & N/A\\
        % Studer et al.~\cite{studerComprehensiveStudyImageNet2019a} & 30 & DeepLabV3     & $\sim$8h    & 92.90 & N/A\\
        % System-5~\cite{simistiraICDAR2017CompetitionLayout2017} & 30 & FCN$^\ast$ & N/A   & 98.35 & N/A\\
        % System-4.1~\cite{simistiraICDAR2017CompetitionLayout2017}& 30 & ResNet18     & N/A    & 98.64 & N/A\\
        \cite{studerComprehensiveStudyImageNet2019a} & 2019 & SegNet  & $\sim$8h    & 86.90\\
        \cite{studerComprehensiveStudyImageNet2019a} & 2019 & DeepLabV3  & $\sim$8h & 92.90\\
        \hline
        Ours & 2023  & SegNet    & $\sim$4.5h & 92.61 \\
        Ours & 2023  & DeepLabv3 & $\sim$3.5h & 93.04 \\
        \hline
    \end{tabular}
\end{table}

The experiments conducted with DIVA-DAF were significantly faster compared to the implementation of Studer et al. 
For both architectures, we achieved similar results.
The performance difference of SegNet~\cite{badrinarayanan2017segnet} is probably due to different default parameters not mentioned by Studer et al.
However, we have time savings for DeepLabv3~\cite{chen2017rethinking} and SegNet of more than 55\% and 45\%, respectively.
We think this is caused by the efficient data loading in DIVA-DAF (from the file system but also into the GPU) and the improvements in driver technology.

\begin{comment}
\begin{table}[h]
  \caption{Implementation Time in minutes and code duplication in lines of code}
  \label{tab:application_implementation}
  \begin{tabular}{lcccc}
    \toprule
    Module part & \multicolumn{2}{c}{Programming time [m]} & \multicolumn{2}{c}{Code Duplication}\\
                & Scratch & DIVA-DAF & Scratch & DIVA-DAF\\
    \midrule
    Data loading            & 150 & 2 & 0 & 0 \\
    Network                 & 30  & 5 & 0 & 0 \\
    Execution               & 10  & 5 & 0 & 0 \\
    \midrule
    S1. Pre-training           & 15 & 2  & 0  & 0 \\
    S2. Comparing network             & 20 & 2  & 30 & 0 \\
    S3. Visualizing             & 25 & 20 & 5 & 0 \\
    \bottomrule
  \end{tabular}
\end{table}
\end{comment}

\section{Conclusion and Future Work}
% introduced an open-source python dl framework that is built for DIA
In this paper, we introduce DIVA-DAF. 
It is an open-source PyTorch-Lightning-based deep learning framework designed to create rapid prototypes and reproducible experiments for the historical document analysis community.

The framework offers pre-implemented tasks that are easily adaptable, including segmentation, classification, and object detection.
%or the possibility to implemented a custom deep learning task. 
It is also possible to implement custom tasks and data modules in a straightforward way due to the framework's abstract classes. % compatibility with other Frameworks
As shown in the application part, DIVA-DAF allows users to gain efficiency during implementation as well as model execution. 

% limitations of the framework
\rev{However, the framework has certain functional limitations like conducting multi-runs within the framework, doing hyperparameter optimization, running tasks with multiple headers or losses, and downloading datasets in an automatic fashion.}
% % can not conduct multiple runs within the framework (bash workaround) ->  Hyperparameter optimization
% % multi header and multi loss
% % automatic dataset download

To encourage a larger public to use the framework, a user interface could be developed.
To improve the framework, we envisage \rev{implementing} new tasks in document analysis, \rev{adding} new networks, \rev{integrating} new ground truth formats, \rev{ and improving its documentation, which this paper is part of}.
To further support the users in analyzing their results, it would be interesting to add classification activation maps and filter visualization techniques to the framework.

% what do we want to add in the future
% To improve the framework, we want to extend our set of tasks with more \ac{DIA} relevant scenarios to make it even more attractive for the community.
% more tasks, especially self-supervised once
% Also, we want to add more pretraining strategies to the framework to tackle the lack of manually-labeled ground truth.
% additional visualization tools like CAM or layerwise activation

% improve the stability and general structure
% Overall, we continue to improve the general structure and efficiency of DIVA-DAF.

% Future work: GUI for humanities

% \begin{acks}
% \end{acks}

% \newpage
\bibliographystyle{ACM-Reference-Format}
\bibliography{bibtex}

\end{document}